%% file: 0_paper.tex
\documentclass[sigconf]{acmart}

\usepackage{algorithm}
\usepackage{algorithmic}
\usepackage{url}
\usepackage{bbm}
\usepackage{bm}
\usepackage{subcaption}
\usepackage[export]{adjustbox}
\usepackage{amsmath}
\usepackage{mathtools}

\AtBeginDocument{%
  \providecommand\BibTeX{{%
    \normalfont B\kern-0.5em{\scshape i\kern-0.25em b}\kern-0.8em\TeX}}}

\setcopyright{acmcopyright}
\copyrightyear{2022}
\acmYear{2022}
\setcopyright{acmcopyright}\acmConference[BCB '22]{13th ACM International
Conference on Bioinformatics, Computational Biology and Health Informatics}{August
7--10, 2022}{Northbrook, IL, USA}
\acmBooktitle{13th ACM International Conference on Bioinformatics, Computational
Biology and Health Informatics (BCB '22), August 7--10, 2022, Northbrook, IL, USA}
\acmPrice{15.00}
\acmDOI{10.1145/3535508.3545516}
\acmISBN{978-1-4503-9386-7/22/08}

\acmSubmissionID{127}

\begin{document}

\title{Mitigating Health Disparities in EHR via Deconfounder}

\author{Zheng Liu}
\email{zliu212@uic.edu}
\affiliation{%
  \institution{University of Illinois at Chicago}
  \city{Chicago}
  \country{USA}}
  
\author{Xiaohan Li}
\email{xli241@uic.edu}
\affiliation{%
  \institution{University of Illinois at Chicago}
  \city{Chicago}
  \country{USA}}

\author{Philip Yu}
\email{psyu@uic.edu}
\affiliation{%
  \institution{University of Illinois at Chicago}
  \city{Chicago}
  \country{USA}}

\renewcommand{\shortauthors}{Liu, et al.}

\begin{abstract}

\input{text/1-Abstract}
\end{abstract}



\maketitle

\input{text/2-Intro}

\input{text/4-Methods}

\input{text/5-Exp}

\input{text/6-RelatedWorks}

\input{text/7-Conclusion}

\bibliographystyle{ACM-Reference-Format}
\bibliography{ref}

\end{document}

%% file: text/1-Abstract.tex
Health disparities, or inequalities between different patient demographics, are becoming a crucial issue in medical decision-making, especially in Electronic Health Record (EHR) predictive modeling. In order to ensure the fairness of sensitive attributes, conventional studies mainly adopt calibration or re-weighting methods to balance the performance on among different demographic groups.
However, we argue that these methods have some limitations. First, these methods usually mean making a trade-off between the model's performance and fairness. Second, 
many methods attribute the existence of unfairness completely to the data collection process, which lacks substantial evidence.
In this paper, we provide an empirical study to discover the possibility of using deconfounder to address the disparity issue in healthcare. Our study can be summarized in two parts. The first part is a pilot study demonstrating the exacerbation of disparity when unobserved confounders exist. The second part proposed a novel framework, Parity Medical Deconfounder (PriMeD), to deal with the disparity issue in healthcare datasets. Inspired by the deconfounder theory, PriMeD adopts a Conditional Variational Autoencoder (CVAE) to learn latent factors (substitute confounders) for observational data, and extensive experiments are provided to show its effectiveness.

%% file: text/2-Intro.tex
\section{Introduction}
Machine learning models have demonstrated the promising potential on Electronic Health Records (EHRs), such as risk prediction \cite{ma2018risk,cheng2016risk}, auxiliary diagnosis \cite{rajkomar2018scalable}, and automated prescription \cite{zhang2017leap}. However, recent evidence shows these models also exacerbate bias and disparities in healthcare, which raises considerable concern and criticism \cite{chen2021algorithm,pfohl2021empirical,chen2019can,chen2020ethical,zhou2021radfusion,mhasawade2021machine,gianfrancesco2018potential}. Many studies have shown that a machine learning model may provide disparate results for people with different backgrounds. For example, for patients of different races, the accuracy and quality of a machine learning model's prediction may vary significantly \cite{obermeyer2019dissecting}. 
This disparity reduces the utility of machine learning models and is especially detrimental to disadvantaged and underrepresented populations \cite{chen2021algorithm,pfohl2021empirical}. 

To deal with this fairness issue, 
conventional studies adopt certain criteria such as 
equalized odds \cite{hardt2016equality}
or counterfactual fairness \cite{kusner2017counterfactual} to mitigate the disparities \cite{pfohl2019counterfactual,kamiran2012data,pleiss2017fairness}. However, we argue that there are limitations to these approaches. 
First, maintaining fairness usually means sacrificing some precision of the model, especially when the model penalizes/calibrates the majority of datapoints \cite{feldman2015certifying,pleiss2017fairness}. If a model's performance deteriorates after adopting certain fairness criteria, the utility of the model does not necessarily improve.
Second, 
many methods attribute the existence of unfairness completely to the data collection process, which lacks substantial evidence. For example, many causality-based fairness methods assumes ``no unobserved confounders'' assumption \cite{dwork2012fairness,kilbertus2020sensitivity}, which usually doesn't hold on EHR data. 

To address these issues, we first focus on the causes of health disparity and construct a fairness model on this basis. Consider a predictive task that using the observation (including both \textbf{sensitive attributes} and other \textbf{clinical observations})  to predict the \textbf{outcomes} of patients. In this paper we attribute the cause of health disparity to two factors: the imbalance of sensitive attributes and the existence of unobserved confounders (e.g., ethnicity, social/economic status, etc.) that affects both clinical observations and outcomes.

Then, we propose a novel framework, Parity Medical Deconfounder (PriMeD), to deal with the above two factors. Inspired by the deconfounder theory \cite{wang2019blessings,wang2020towards} and the Inverse Propensity Weighting (IPW) methods \cite{haukoos2015propensity}, PriMeD can provide accurate and fair prediction by addressing the above two factors of health disparity.
In detail, PriMeD is a two-stage supervised framework that infers unobserved confounders in the first stage and makes predictions in the second stage. 
In the first stage, PriMeD resorts to a weighted Conditional Variational Auto-encoder (CVAE) \cite{sohn2015learning} to learn fair latent representation for observational data, and the weight of each datapoint is the probability of the sensitive attributes exist. Therefore, the imbalance of sensitive attributes is addressed in this step, and according to the deconfounder theory, we regard the fair representation as substitute confounders. 
In the second stage, PriMeD adopts a self-attentive deep neural network to predict medical outcomes based on the clinical observations, sensitive attributes and the fair representation of the first stage. 
The goal of this step is to achieve higher accuracy in prediction, and theoretically any predictive model is eligible here. In this paper, we adopt an attention neural network to achieve this goal. 

In summary, in this paper our contribution is listed as follows:

\begin{itemize}
    \item To the best of our knowledge, this paper is the first one using deconfounder theory to address health disparities in EHR analysis.
    \item We propose two causes of health disparity and design a model based on this judgement. Parity Medical Deconfounder (PriMeD), is the novel model we proposed to address the health disparity in EHR modeling.
    \item Extensive experiments are provided to show the superiority of PriMeD in achieving equalized odds\cite{hardt2016equality} and making accurate decisions.
\end{itemize}

%% file: text/4-Methods.tex
\begin{figure}
    \begin{subfigure}[]{0.45\textwidth}
        \centering
        \includegraphics[height=0.6in]{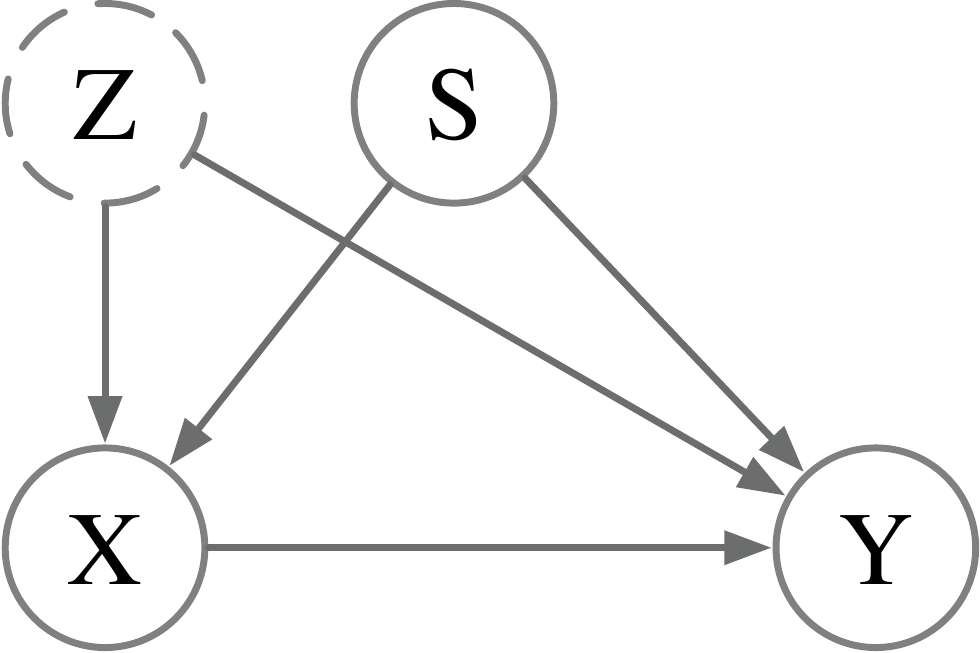}
        \caption{The SCM of EHR. Circle with dotted line denotes unobserved variable.} 
        \label{fig:ovw}
    \end{subfigure}
    \begin{subfigure}[]{0.5\textwidth}
        \centering
        \includegraphics[height=2in]{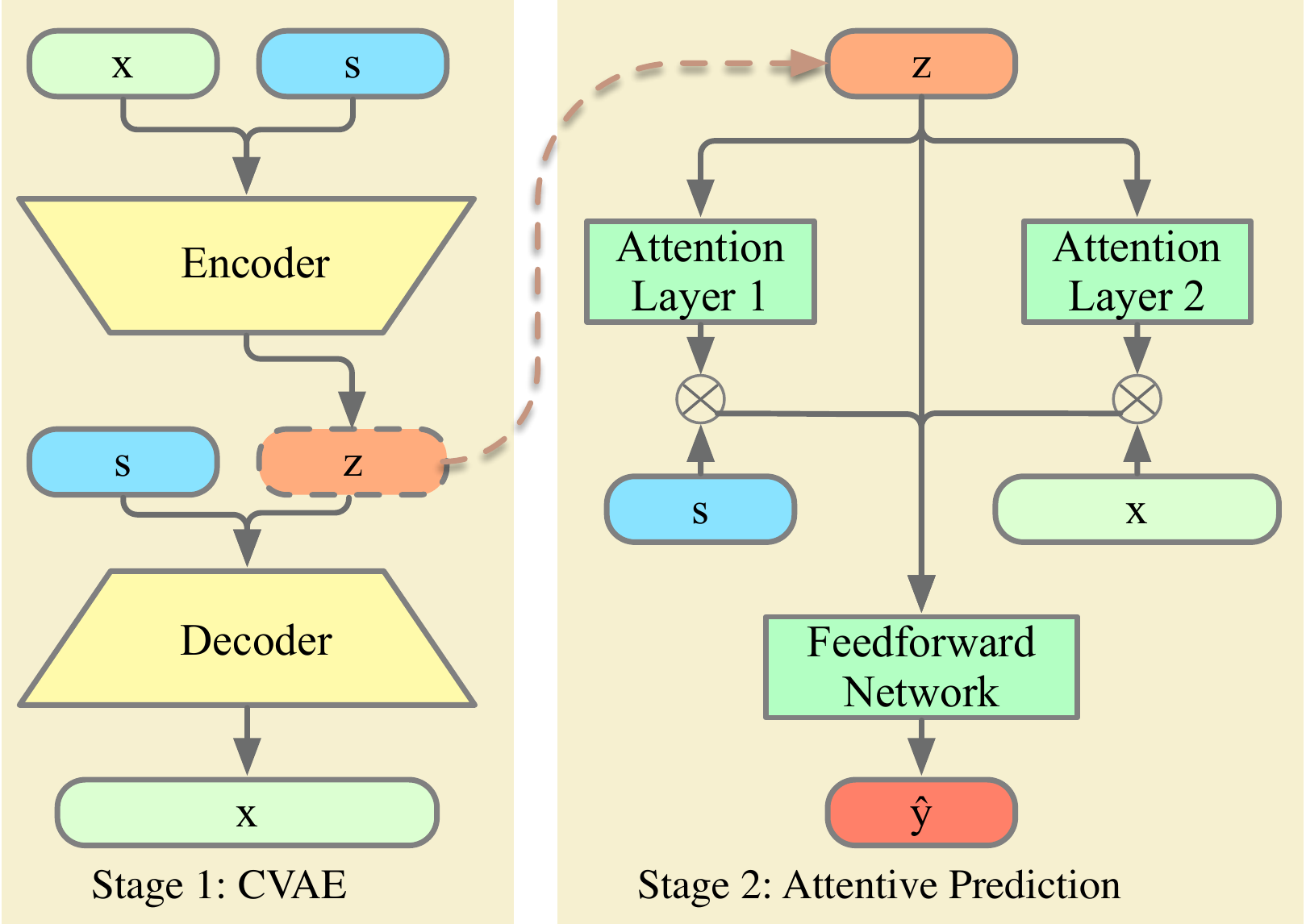}
        \caption{The architecture of PriMeD.} 
        \label{fig:ovw2}
    \end{subfigure}
    \caption{The causal graph and model architecture of PriMeD. $X$, $Y$, $S$ and $Z$ denote clinical features, outcomes, sensitive attributes and confounders. Rounded rectangles ($x$, $\hat{y}$, $b$, $z$) in Figure \ref{fig:ovw2} denote corresponding data vector.} 
    \label{fig:demo}
\end{figure}

\section{Method}

\subsection{Problem Definition}
We consider an EHR with the latent structure shown in Figure \ref{fig:ovw}. In this causal diagram, an EHR consists of three parts: the sensitive attributes $S$ (e.g., gender, race, etc.), the clinical observations $X$ and the outcome/predictor $Y$.
We also assume the existence of unobserved confounders $Z$ (e.g., living habits, social/economic status, genotypes, etc.) in the diagram. Please note that although $Z$ may be associated with $B$ in some scenarios, in this paper we don't consider this effect.

In practice, we have an EHR dataset $\mathcal{D} = \{ (\bm x_i, \bm s_i, y_i) \}_{i=1}^N$ where $\bm x$, $\bm s$ and $y$ corresponds to $X$, $S$, and $Y$ in Figure \ref{fig:ovw}. Our goal is using $\bm x$ and $\bm s$ to predict $y$, while minimizing the the gap between performances on different subgroups.

\subsection{PriMeD}
Inspired by the deconfounder theory, PriMeD consists of two stages. The first stage is to address the disparity by learning fair representations, and the second stage is to make accurate predictions.

\paragraph{\textbf{Stage 1: weighted Conditional Variational Auto-encoder (CVAE)}}
In the first stage, PriMeD addresses the confounding effect of $Z$ and $S$ by learning fair latent representations. Here we adopt a weighted Conditional Variational Auto-encoder (CVAE) \cite{sohn2015learning} to achieve this goal. 
As a variant of Variational Autoencoder (VAE) \cite{kingma2013auto}, CVAE regards the observations as consequences of a latent factor and a condition. In EHR modeling, the clinical observations $X$ are the effect of latent factor $Z$ and sensitive attributes $S$, which is identical to the CVAE setting.

CVAE introduces an variational posterior $q_\phi(\bm z|\bm x, \bm s)$ as a recognition network, to deal with the intractability of true posterior $p_\theta(\bm z|\bm x, \bm s)$ in maximum likelihood inference. It exhibits an autoencoder structure, with the 
recognition network $q_\phi(\bm z|\bm x, \bm s)$ as encoder and the generation network $p_\theta(\bm x|\bm z,\bm s)$ as decoder. 
When optimizing CVAE, the objective is designed to maximize the variational lower bound (ELBO) of the log-likelihood, which is written as
\begin{equation}
    \begin{aligned}
    \log p_\theta(\bm x|\bm s)  \ge &
    - KL( q_\phi(\bm z|\bm x, \bm s) || p_\theta(\bm z) ) \\
    & + \mathbb{E}_{q_\phi(\bm z|\bm x, \bm s)}[\log p_\theta(\bm x|\bm z,\bm s)].
    \end{aligned}
    \label{eq:cvae}
\end{equation}

To approximate the second term, we draw $L$ samples $\bm z^{(l)} (l=1, \cdots, L)$ from the recognition distribution $q_\phi(\bm z|\bm x, \bm s)$, and rewrite the empirical objective for datapoint $(\bm x, \bm s)$ as

\begin{equation}
    \begin{aligned}
    \mathcal{L}(\bm x, \bm s; \theta, \phi)  = &
    - KL( q_\phi(\bm z|\bm x, \bm s) || p_\theta(\bm z) ) \\
    & + \frac{1}{L} \sum_{l=1}^{L} \log p_\theta(\bm x|\bm z^{(l)},\bm s)
    \end{aligned}
    \label{eq:cvae}
\end{equation}

where the first term minimizes the difference between the posterior and the prior of $\bm z$, and the second term minimizes the difference between the input and output of CVAE. In this paper, we set $p_\theta(\bm z)$ as standard Gaussian distribution. By adopting such an architecture, the confounding effect of $X$ and $S$ can be addressed.

Moreover, to deal with the imbalanced distribution of $S$, in the training process we assign a weight $\omega_{\bm s}$ for each datapoint $(\bm x, \bm s)$ in $\mathcal{D}$.
The intuition of using the weight is straightforward: 
since the distribution of sensitive attributes $\bm s$ in the dataset is usually imbalanced, we would like to assign larger weights for rarer $\bm s$ to mitigate this imbalance.

To obtain $\omega_{\bm s}$, suppose there are $J$ sensitive attributes $\bm s = [s_1, \cdots, s_J]$ for each patient in the dataset. The training weight $\omega_{\bm s}$ for patient with sensitive attributes $\bm s$ is defined as:
\begin{equation}
    \omega_{\bm s} = \frac{1}{ \prod_{j=1}^J freq(s_j) }
\end{equation}

where $freq(s_j)$ is the frequency of certain attribute that occurs in the dataset. Hence, the more frequent one attribute occurs in the sensitive attribute vector $\bm s$, the smaller the corresponding  $\omega_{\bm s}$ is. Since we usually won't focus on too many sensitive attributes, such an approximation of propensity score is enough to mitigate the imbalance of data distribution. The objective for the whole dataset can be written as

\begin{equation}
    \mathcal{L}_{CVAE}( \mathcal{D} ; \theta, \phi) =  \sum_{i=1}^{N} \omega_{\bm s} \mathcal{L}(\bm x_i, \bm s_i; \theta, \phi)
\end{equation}

where $(\bm x_i, \bm s_i)$ is $\bm x$ and $\bm s$ of the $i$-th datapoint in dataset $\mathcal{D}$.

\paragraph{\textbf{Stage 2: Prediction with Attention}  }
Once the CVAE is trained successfully, the learned latent variable $\bm z$ can be viewed as substitute confounders and be used to derive accurate and unbiased predictions.
Theoretically, any model taking $\bm x$, $\bm s$ and $z$ as inputs is eligible in this stage. Here we use a deep learning model with attention layers to achieve higher prediction performance. Suppose the length of sensitive attributes $\bm z$ is $K$, and
the clinical features $\bm x = [x_1,x_2,\cdots,x_M]$ is with $M$ digits. We use $\bm z$ to learn attention weights $\bm w_{\bm x} = [w_1,w_2,\cdots,w_M]$ in order to determine which digit of $\bm x$ is more important in prediction.
Formally, PriMeD learns $\bm w_{\bm x}$ with a simple attention module as follows:

\begin{equation}
    \bm w_{\bm x} = softmax(\bm W_{x}^{att}\bm z^\intercal)
\end{equation}
where $\bm W_{\bm x}^{att}$ is the $M \times K$ parameter matrix and $softmax(\cdot)$ is the softmax function. 
Similarly, we also have $\bm w_{\bm s} = softmax(\bm W_{\bm s}^{att} \bm z^\intercal)$ to learn attention weights for sensitive attributes $\bm s$.
Finally, the processed features are input into a feed-forward neural network to derive the final decision:

\begin{equation}
    \hat{y} = MLP ([\bm x \odot \bm w_x^\intercal;\bm s \odot \bm w_s^\intercal; \bm z]),
\end{equation}

where $[; ]$ denotes the concatenation operation, $\odot$ denotes the element-wise multiplication operation, and $MLP (\cdot)$ is a two-layer MLP. In this stage, by incorporating the latent factor $\bm z$ in the prediction, the model can calibrate the bias from observational data and derive a higher-quality prediction:

%% file: text/5-Exp.tex
\begin{table}[]
\caption{Information used in our experiments.}
\label{tab:exp_info}
\begin{tabular}{c|cc}
\hline Datasets & MIMIC & Hip \& Knee \\ \hline
\begin{tabular}[c]{@{}c@{}}Sensitive\\ Attributes\end{tabular}      & \begin{tabular}[c]{@{}c@{}}
Insurance, Ethnicity\\ Gender

\end{tabular} & \begin{tabular}[c]{@{}c@{}}
Gender, Race, Age

\end{tabular}  

\\ \hline
\begin{tabular}[c]{@{}c@{}}Clinical \\ Features\end{tabular} & \begin{tabular}[c]{@{}c@{}}

Demographics\\ Clinical events, Lab events\\ Procedures, Diagnoses\end{tabular}                               & \begin{tabular}[c]{@{}c@{}}

Height, Weight\\
Position \# \\ 20-37,  51-63

\end{tabular} \\ \hline
Outcomes                                                     & Mortality                                                                                                                                              & \begin{tabular}[c]{@{}c@{}}Re-operation\\ Re-admission\\ DOptoDis\end{tabular}                                                                                              \\ \hline
\end{tabular}
\end{table}

\section{Experiments}
In this section, we conduct experiments to evaluate the performance of our proposed PriMeD.

\subsection{Datasets}

We use three real-world EHR datasets to measure PriMeD's performance.

\begin{itemize}
    
    \item \textbf{\textit{MIMIC-III \footnote{https://physionet.org/content/mimiciii/1.4/}}}. MIMIC-III\cite{johnson2016mimic} is a deidentified, publicly-available dataset comprising comprehensive clinical data of patients admitted to the Beth Israel Deaconess Medical Center. MIMIC-III contains EHRs associated with 46,520 patients, including over 20 tables such as medical events, diagnoses, prescriptions, etc.
    
    \item \textbf{\textit{Hip \& Knee}}. 
    Hip \& Knee are two subsets of the National Surgical Quality Improvement Program (NSQIP) project\footnote{https://www.facs.org/quality-programs/data-and-registries/acs-nsqip/} \cite{cohen2016improved}. NSQIP provides clinical observations and observations for patients taking surgical operations to track surgical complications after operations. 
    They are associated with patients receiving 
    Hip Arthroplasty (CPT\footnote{https://medicaid.ncdhhs.gov/blog/2021/12/30/cpt-code-update-2022} 27130) 
    and
    Knee Arthroplasty (CPT 27447) surgical operations. 
    There are 96,441 and 156,292 pieces of data in them respectively, and each data point contains demographics, pre-operative features, and the outcome of the surgery for a patient.

\end{itemize}

\paragraph{\textbf{data preprocessing}}
According to the documentation of MIMIC-III\footnote{https://mimic.mit.edu/docs/iii/tables/} and NSQIP\footnote{https://www.facs.org/quality-programs/data-and-registries/acs-nsqip/participant-use-data-file/}, we determine the columns/tables used in our experiments (shown in Table \ref{tab:exp_info}).
In the MIMIC-III dataset, we regard the patient's insurance status, ethnicity and gender as sensitive attributes, and use clinical events to predict mortality. 
In the Hip \& Knee dataset, we take gender, race and age as sensitive attributes and use BMI (inferred from height and weight) and multiple clinical observations to predict the readmission, reoperation, and whether there is a prolonged hospitalization after the operation (DOptoDis > 5).

\subsection{Evaluation Metrics \& Implementation Details}
Since all tasks are binary classifications, we use Area Under the Receiver Operating Characteristic curve (AUROC) to measure its accuracy in classification. 
We implement the model using Pytorch 1.10 and adopt the $10^{-4}$ as the learning rate and $5 \times 10^{-4}$ as the weight decay. The ratio of training, validating, and testing set is 7:2:1. 

\begin{table*}[]
\caption{Accuracy of PriMeD and other baselines in prediction.}
\begin{tabular}{c|c|ccc|ccc}
\hline
\textbf{Dataset} & MIMIC-III       & \multicolumn{3}{c|}{Knee}                                                                                & \multicolumn{3}{c}{Hip}                                                                                  \\ \hline
\textbf{Task}    & Mortality       & Reoperation     & Readmission     & \begin{tabular}[c]{@{}c@{}}Prolonged \\ Hospitalization\end{tabular} & Reoperation     & Readmission     & \begin{tabular}[c]{@{}c@{}}Prolonged \\ Hospitalization\end{tabular} \\ \hline
DNN              & 0.6984          & 0.6752          & 0.6342          & 0.7061                                                               & 0.7044          & 0.6448          & 0.7403                                                               \\
Re-weighting     & 0.6553          & 0.6457          & 0.5973          & 0.6590                                                               & 0.6793          & 0.6125          & 0.6983                                                               \\
CE odds          & 0.6810          & 0.6579          & 0.6192          & 0.6773                                                               & 0.6889          & 0.6313          & 0.7045                                                               \\
WFC              & 0.6741          & 0.6481          & 0.6008          & 0.6717                                                               & 0.6865          & 0.6217          & 0.7102                                                               \\
RFC              & 0.6889          & 0.6630          & 0.6289          & 0.6928                                                               & 0.6956          & 0.6375          & 0.7363                                                               \\
FuCS             & 0.6944          & 0.6774          & 0.6375          & 0.7024                                                               & 0.7029          & 0.6426          & 0.7345                                                               \\ \hline
\textbf{PriMeD}  & \textbf{0.7013} & \textbf{0.6821} & \textbf{0.6413} & \textbf{0.7146}                                                      & \textbf{0.7059} & \textbf{0.6453} & \textbf{0.7457}                                                      \\ \hline
\end{tabular}
\label{tab:perform}
\end{table*}

\subsection{Comparison Experiments}

In this subsection, we compare our model with several baseline models listed below.

\begin{itemize}
    \item \textbf{DNN}. We adopt a deep neural network to make accurate predictions on observational data. This method is served as a baseline to show the original disparity without interference.

    \item \textbf{Re-weighting}\cite{kamiran2012data}. Kamiran et al.   propose a pre-processing method based on the dataset re-weighting to remove bias from the dataset. Re-weighting can reduce the discrimination while maintaining the overall positive class probability for the training set. We use this method to assign weights for the data, while using the same model as \textbf{DNN} to make the prediction.

    \item \textbf{CE Odds}\cite{pleiss2017fairness}. Pleiss et al.  propose a post-processing method based on calibration constraints to minimize error disparity while maintaining calibrated probability estimates.

    \item \textbf{WFC}\cite{jiang2020wasserstein}. Wasserstein Fair Classification is a post-processing method that enforces independence between the classifier outputs and sensitive information by minimizing Wasserstein-1 distances. 
    
    \item \textbf{RFC}\cite{agarwal2018reductions}. Agarwal et al. propose an in-processing method, focusing on reducing fair classification to a sequence of cost-sensitive classification problems to achieve the lowest error subject to the desired constraints.
    
    \item \textbf{FuCS}\cite{rezaei2021robust}. Rezaei et al. propose a pre-processing method to guarantee fairness under covariate shift in which the covariates change while the conditional label distribution remains the same.

\end{itemize}

\subsection{Performance of Prediction}
Table \ref{tab:perform} shows the performance of all baselines together with PriMeD on all three tasks. From the table, we can observe that PriMeD outperforms all other baselines significantly in AUROC. It is because the architecture of PriMeD is specially designed for EHRs, while other baselines are universal classifiers. PriMeD also outperforms the vanilla classifier, DNN, on all tasks, showing its superiority in modeling EHR.
Apart from PriMeD, FuCS achieves the second-best performance, which may due to its ability to deal with non-iid data. When the confounder is unobserved, EHR data may exhibit multiple distributions and is not iid. FuCS also outperforms DNN on some tasks, indicating its ability to learn knowledge from non-iid data.
Then, RFC and CE odds achieve lower performance at the price of guaranteeing equalized odds in prediction. WFC and Re-weighting method have the worst performances. This fact indicates their means of controlling fairness clearly prevent them from learning from observational data. 

If we compare different datasets/tasks, we will find the difficulty of tasks is different as well. As for the reoperation task, the differences between the performances of methods are relatively small, indicating the confounding effect on this predictor is relatively weaker. The difference between methods on the prolonged hospitalization task is quite large, indicating this predictor is strongly influenced by the confounding effect.

\begin{table*}[]
\caption{The extent of disparity of PriMeD and other baselines in prediction.}
\label{tab:disparity}
\begin{tabular}{c|cc|cc|cc}
\hline
Dataset      & \multicolumn{2}{c|}{\begin{tabular}[c]{@{}c@{}}MIMIC-III \\ Mortality\end{tabular}} & \multicolumn{2}{c|}{\begin{tabular}[c]{@{}c@{}}Knee\\ Prolonged Hospitalization\end{tabular}} & \multicolumn{2}{c}{\begin{tabular}[c]{@{}c@{}}Hip\\ Prolonged Hospitalization\end{tabular}} \\ \hline
Sensitive Attribute      & Insurance & Ethnicity  &  Race & Age  & Race  & Age      \\ \hline
DNN          & $1.570\times10^{-2}$  & $1.802\times10^{-2}$ & $9.618\times10^{-3}$ & $10.859\times10^{-3}$  & $8.702\times10^{-3}$  & $9.859\times10^{-3}$ \\
Re-weighting & $1.378\times10^{-2}$  & $1.638\times10^{-2}$ & $8.637\times10^{-3}$ & $10.178\times10^{-3}$ & $8.418\times10^{-3}$ & $9.163\times10^{-3}$ \\
CE odds      & $1.384\times10^{-2}$  & $1.667\times10^{-2}$ & $8.825\times10^{-3}$ & $9.074\times10^{-3}$ & $8.258\times10^{-3}$  & $8.741\times10^{-3}$ \\
WFC          & $1.463\times10^{-2}$  & $1.612\times10^{-2}$ & $8.363\times10^{-3}$ & $8.523\times10^{-3}$   & $8.357\times10^{-3}$  & $8.275\times10^{-3}$ \\
RFC          & $1.249\times10^{-2}$  & $1.563\times10^{-2}$ & $7.356\times10^{-3}$ & $7.804\times10^{-3}$   & $7.304\times10^{-3}$ & $8.002\times10^{-3}$  \\
FuCS         & $1.120\times10^{-2}$  & $1.496\times10^{-2}$ & $7.576\times10^{-3}$ & $7.063\times10^{-3}$ & $7.776\times10^{-3}$ & $8.024\times10^{-3}$ \\\hline
PriMeD       & $1.123\times10^{-2}$  & $1.470\times10^{-2}$ & $6.982\times10^{-3}$ & $9.451\times10^{-3}$  & $7.227\times10^{-3}$ & $7.769\times10^{-3}$                                     \\  \hline
\end{tabular}
\end{table*}

\subsection{Mitigate Health Disparity}
In this subsection, we demonstrate the disparity of prediction between different subgroups. 
Table \ref{tab:disparity} demonstrates the extent of disparities in prediction with respect to different datasets, predictors and baseline models. In this table, the Insurance column shows the difference in AUROC between patients with public and private insurances, the Ethnicity and Race columns show the difference between White and non-White patients, and the Age column shows the difference between patients with ages above and below 65.

From Table \ref{tab:disparity}, we can observe that PriMeD can effectively reduce the difference between subgroups in prediction. It achieves the fairest performance across all datasets and all metrics. Apart from PriMeD, we can observe that FuCS and RFC achieve the second-best performance, which means they can balance the performance on different demographic subgroups. From the table, we also observe that the disparity in age is more stubborn than the disparity of race, which means age is a stronger confounder that can cause bias.

\begin{table}[]
\caption{The performance of different variants of PreMeD.}
\label{tab:ablation}
\begin{tabular}{c|cc|cc}
\hline
               & \multicolumn{2}{c|}{\begin{tabular}[c]{@{}c@{}}Knee\\ Reoperation\end{tabular}} & \multicolumn{2}{c}{\begin{tabular}[c]{@{}c@{}}Knee\\ Readmission\end{tabular}} \\ \hline
               & AUROC                                 & Diff\_Race                               & AUROC                                & Diff\_Race                               \\ \hline
PriMeD         & 0.6821                                & 0.005463                                  & 0.6413                               & 0.005731                               \\
PriMeD-Stage 1 & 0.6410                                & 0.005585                                & 0.5903                               & 0.006121                                \\
PriMeD-Stage 2 & 0.6748                                & 0.008463                                & 0.6370                               & 0.007951                              \\ \hline
\end{tabular}
\label{aba}
\end{table}

\subsection{Ablation Study}
In order to show the effectiveness of each stage of PriMeD, Table \ref{aba} demonstrates the performance of each module of PriMeD. In this module, PriMeD-Stage 1 means the performance of prediction while only using the stage 1 model (the CVAE). In PriMeD-Stage 1 we use the latent vector $z$ the CVAE learned to make classifications. Similarly, PriMeD-Stage 2 means simply using a deep neural network to make the prediction. In PriMeD-Stage 2, we use the input vector $b$ and $x$ to substitute the position of $z$ in the stage 2 model. In Table \ref{aba}, we use AUROC to describe the performance in prediction and use the difference between White and non-White patients (denoted by Diff\_Race) to measure the health disparity. We can observe that the CVAE can learn relatively fair predictions while the accuracy of classification is low. The attention neural network can achieve quite an accurate prediction which is similar to the DNN baseline, while the disparity issue is unaddressed.

%% file: text/6-RelatedWorks.tex
\section{Related Works}
The related works of this paper consist of three parts: the observation of healthcare disparity, the review of fairness in machine learning, and a brief introduction of the deconfounder theory.
\subsection{Healthcare Disparity}
With the increase in the number of machine learning applications in the healthcare domain, more concerns have been raised regarding the potential ethical issue of these models \cite{pfohl2021empirical,chen2020ethical,chen2021algorithm}. Recently, many studies have shown that these models worsen existing health disparities. Health disparity means the inequity for different patient subgroups when providing healthcare services \cite{carter2002health,fikree2004role}. Although this inequity 
naturally exists in observational data \cite{krasanakis2018adaptive,obermeyer2019dissecting}, it is amplified by the algorithmic bias in the model \cite{mhasawade2021machine,mhasawade2021machine}. In response, considerable attention has been devoted to dissecting the disparities in the machine learning models \cite{zhou2021radfusion,seyyed2020chexclusion}. According to the recent analysis, 
this disparity can result from the inequity in care access and delivery history \cite{chen2020treating,gaskin2012residential}, the existence of underrepresented groups in clinical datasets \cite{larrazabal2020gender}, the misuse of biased features during model development \cite{kallus2018residual,jiang2020identifying}, and difference of distribution between training and teat sets
\cite{veinot2018good}.

\subsection{Fairness \& Algorithmic Debiasing}
Traditionally, fairness avoids any prejudice or favoritism towards any individual or group in the decision-making process.
Based on this principle, many fairness metrics are proposed and can be categorized mainly into three classes. 
One standard fairness metric is group fairness, such as demographic parity, which requires the probability of a positive prediction to be the same for each group \cite{zemel2013learning}. Another fairness metric is individual fairness, such as counterfactual fairness. It requires  the probability of a positive prediction to be the same for both factual and counterfactual datapoints \cite{kusner2017counterfactual}. In this paper, our goal is to minimize the difference in utility between patient subgroups, which is similar to  \cite{rajkomar2018ensuring,heidari2019moral,pfohl2019counterfactual} and is different from either group or individual fairness.

Based on the above definitions of fairness, there are three types of algorithms to address the fairness of machine learning models: the pre-processing, in-processing, and post-processing methods.
Pre-processing methods either change the label of training data or assign weights for data before the training process \cite{kamiran2012data,kamiran2010classification}. 
Post-processing methods conduct calibration of prediction results after training. Some of them calibrate the prediction based on a holdout set, and others are based on certain constraints to ensure fairness \cite{d2017conscientious,pleiss2017fairness}.
In-processing techniques develop model architectures to remove discrimination during the model training process \cite{d2017conscientious,louizos2015variational,sattigeri2019fairness}. Our model adopts the in-processing technique, because it is more flexible in dealing with complex latent data structures.

\subsection{Causal Inference with Unobserved Confounders}
In this part, we mainly discuss the deconfounder theory. 
Deconfounder \cite{wang2019blessings,wang2020towards} is a theory to estimate unbiased treatment effects for observational data with the setting of multiple causes. Due to unobserved confounders of many tasks, traditional methods can hardly learn unbiased knowledge from the observational data. However, suppose we can observe multiple causes of the outcome. In this case, 
the dependencies between causes can be used to infer latent variables. The latent variables can be used as substitutes for the hidden confounders. Therefore, Wang \& Blei propose a two-stage architecture to deal with the unobserved confounders. Recently, deconfounder has been applied to many areas such as recommender systems \cite{wang2020causal} and medical treatment estimation \cite{zhang2019medical}. This paper demonstrates that deconfounder can also be used in the fairness domain to reduce inequity in prediction.

%% file: text/7-Conclusion.tex
\section{Conclusion \& Future Directions}

This paper analyzes the disparity between different demographic groups and proposes a solution to address it. Our goal is to minimize the disparity of different demographics while maintaining a high utility for machine learning healthcare applications. To achieve this goal, we propose our PriMeD to derive unbiased predictions. By incorporating CVAE as a module to infer latent factors for patients, PriMeD can naturally correct the disparity in the first stage of the model, while preserving high prediction accuracy. Experiments conducted on three real-world datasets have shown the superiority of PriMeD over other baselines, and the visualization of the learned latent factor further demonstrates its ability to learn fair representation without the influence of sensitive attributes.


\section{Acknowledgement}
This work is supported in part by NSF under grants III-1763325, III-1909323,  III-2106758, and SaTC-1930941. 

%% file: 0_paper.bbl

\begin{thebibliography}{49}


\ifx \showCODEN    \undefined \def \showCODEN     #1{\unskip}     \fi
\ifx \showDOI      \undefined \def \showDOI       #1{#1}\fi
\ifx \showISBNx    \undefined \def \showISBNx     #1{\unskip}     \fi
\ifx \showISBNxiii \undefined \def \showISBNxiii  #1{\unskip}     \fi
\ifx \showISSN     \undefined \def \showISSN      #1{\unskip}     \fi
\ifx \showLCCN     \undefined \def \showLCCN      #1{\unskip}     \fi
\ifx \shownote     \undefined \def \shownote      #1{#1}          \fi
\ifx \showarticletitle \undefined \def \showarticletitle #1{#1}   \fi
\ifx \showURL      \undefined \def \showURL       {\relax}        \fi
\providecommand\bibfield[2]{#2}
\providecommand\bibinfo[2]{#2}
\providecommand\natexlab[1]{#1}
\providecommand\showeprint[2][]{arXiv:#2}

\bibitem[Agarwal et~al\mbox{.}(2018)]%
        {agarwal2018reductions}
\bibfield{author}{\bibinfo{person}{Alekh Agarwal}, \bibinfo{person}{Alina
  Beygelzimer}, \bibinfo{person}{Miroslav Dud{\'\i}k}, \bibinfo{person}{John
  Langford}, {and} \bibinfo{person}{Hanna Wallach}.}
  \bibinfo{year}{2018}\natexlab{}.
\newblock \showarticletitle{A reductions approach to fair classification}. In
  \bibinfo{booktitle}{\emph{International Conference on Machine Learning}}.
  PMLR, \bibinfo{pages}{60--69}.
\newblock


\bibitem[Carter-Pokras and Baquet(2002)]%
        {carter2002health}
\bibfield{author}{\bibinfo{person}{Olivia Carter-Pokras} {and}
  \bibinfo{person}{Claudia Baquet}.} \bibinfo{year}{2002}\natexlab{}.
\newblock \showarticletitle{What is a" health disparity"?}
\newblock \bibinfo{journal}{\emph{Public health reports}}
  \bibinfo{volume}{117}, \bibinfo{number}{5} (\bibinfo{year}{2002}),
  \bibinfo{pages}{426}.
\newblock


\bibitem[Chen et~al\mbox{.}(2020a)]%
        {chen2020treating}
\bibfield{author}{\bibinfo{person}{Irene~Y Chen}, \bibinfo{person}{Shalmali
  Joshi}, {and} \bibinfo{person}{Marzyeh Ghassemi}.}
  \bibinfo{year}{2020}\natexlab{a}.
\newblock \showarticletitle{Treating health disparities with artificial
  intelligence}.
\newblock \bibinfo{journal}{\emph{Nature medicine}} \bibinfo{volume}{26},
  \bibinfo{number}{1} (\bibinfo{year}{2020}), \bibinfo{pages}{16--17}.
\newblock


\bibitem[Chen et~al\mbox{.}(2020b)]%
        {chen2020ethical}
\bibfield{author}{\bibinfo{person}{Irene~Y Chen}, \bibinfo{person}{Emma
  Pierson}, \bibinfo{person}{Sherri Rose}, \bibinfo{person}{Shalmali Joshi},
  \bibinfo{person}{Kadija Ferryman}, {and} \bibinfo{person}{Marzyeh Ghassemi}.}
  \bibinfo{year}{2020}\natexlab{b}.
\newblock \showarticletitle{Ethical Machine Learning in Healthcare}.
\newblock \bibinfo{journal}{\emph{Annual Review of Biomedical Data Science}}
  \bibinfo{volume}{4} (\bibinfo{year}{2020}).
\newblock


\bibitem[Chen et~al\mbox{.}(2019)]%
        {chen2019can}
\bibfield{author}{\bibinfo{person}{Irene~Y Chen}, \bibinfo{person}{Peter
  Szolovits}, {and} \bibinfo{person}{Marzyeh Ghassemi}.}
  \bibinfo{year}{2019}\natexlab{}.
\newblock \showarticletitle{Can AI help reduce disparities in general medical
  and mental health care?}
\newblock \bibinfo{journal}{\emph{AMA journal of ethics}} \bibinfo{volume}{21},
  \bibinfo{number}{2} (\bibinfo{year}{2019}), \bibinfo{pages}{167--179}.
\newblock


\bibitem[Chen et~al\mbox{.}(2021)]%
        {chen2021algorithm}
\bibfield{author}{\bibinfo{person}{Richard~J Chen}, \bibinfo{person}{Tiffany~Y
  Chen}, \bibinfo{person}{Jana Lipkova}, \bibinfo{person}{Judy~J Wang},
  \bibinfo{person}{Drew~FK Williamson}, \bibinfo{person}{Ming~Y Lu},
  \bibinfo{person}{Sharifa Sahai}, {and} \bibinfo{person}{Faisal Mahmood}.}
  \bibinfo{year}{2021}\natexlab{}.
\newblock \showarticletitle{Algorithm fairness in ai for medicine and
  healthcare}.
\newblock \bibinfo{journal}{\emph{arXiv preprint arXiv:2110.00603}}
  (\bibinfo{year}{2021}).
\newblock


\bibitem[Cheng et~al\mbox{.}(2016)]%
        {cheng2016risk}
\bibfield{author}{\bibinfo{person}{Yu Cheng}, \bibinfo{person}{Fei Wang},
  \bibinfo{person}{Ping Zhang}, {and} \bibinfo{person}{Jianying Hu}.}
  \bibinfo{year}{2016}\natexlab{}.
\newblock \showarticletitle{Risk prediction with electronic health records: A
  deep learning approach}. In \bibinfo{booktitle}{\emph{Proceedings of the 2016
  SIAM International Conference on Data Mining}}. SIAM,
  \bibinfo{pages}{432--440}.
\newblock


\bibitem[Cohen et~al\mbox{.}(2016)]%
        {cohen2016improved}
\bibfield{author}{\bibinfo{person}{Mark~E Cohen}, \bibinfo{person}{Yaoming
  Liu}, \bibinfo{person}{Clifford~Y Ko}, {and} \bibinfo{person}{Bruce~L Hall}.}
  \bibinfo{year}{2016}\natexlab{}.
\newblock \showarticletitle{Improved surgical outcomes for ACS NSQIP hospitals
  over time}.
\newblock \bibinfo{journal}{\emph{Annals of surgery}} \bibinfo{volume}{263},
  \bibinfo{number}{2} (\bibinfo{year}{2016}), \bibinfo{pages}{267--273}.
\newblock


\bibitem[d'Alessandro et~al\mbox{.}(2017)]%
        {d2017conscientious}
\bibfield{author}{\bibinfo{person}{Brian d'Alessandro}, \bibinfo{person}{Cathy
  O'Neil}, {and} \bibinfo{person}{Tom LaGatta}.}
  \bibinfo{year}{2017}\natexlab{}.
\newblock \showarticletitle{Conscientious classification: A data scientist's
  guide to discrimination-aware classification}.
\newblock \bibinfo{journal}{\emph{Big data}} \bibinfo{volume}{5},
  \bibinfo{number}{2} (\bibinfo{year}{2017}), \bibinfo{pages}{120--134}.
\newblock


\bibitem[Dwork et~al\mbox{.}(2012)]%
        {dwork2012fairness}
\bibfield{author}{\bibinfo{person}{Cynthia Dwork}, \bibinfo{person}{Moritz
  Hardt}, \bibinfo{person}{Toniann Pitassi}, \bibinfo{person}{Omer Reingold},
  {and} \bibinfo{person}{Richard Zemel}.} \bibinfo{year}{2012}\natexlab{}.
\newblock \showarticletitle{Fairness through awareness}. In
  \bibinfo{booktitle}{\emph{Proceedings of the 3rd innovations in theoretical
  computer science conference}}. \bibinfo{pages}{214--226}.
\newblock


\bibitem[Feldman et~al\mbox{.}(2015)]%
        {feldman2015certifying}
\bibfield{author}{\bibinfo{person}{Michael Feldman}, \bibinfo{person}{Sorelle~A
  Friedler}, \bibinfo{person}{John Moeller}, \bibinfo{person}{Carlos
  Scheidegger}, {and} \bibinfo{person}{Suresh Venkatasubramanian}.}
  \bibinfo{year}{2015}\natexlab{}.
\newblock \showarticletitle{Certifying and removing disparate impact}. In
  \bibinfo{booktitle}{\emph{proceedings of the 21th ACM SIGKDD international
  conference on knowledge discovery and data mining}}.
  \bibinfo{pages}{259--268}.
\newblock


\bibitem[Fikree and Pasha(2004)]%
        {fikree2004role}
\bibfield{author}{\bibinfo{person}{Fariyal~F Fikree} {and}
  \bibinfo{person}{Omrana Pasha}.} \bibinfo{year}{2004}\natexlab{}.
\newblock \showarticletitle{Role of gender in health disparity: the South Asian
  context}.
\newblock \bibinfo{journal}{\emph{Bmj}} \bibinfo{volume}{328},
  \bibinfo{number}{7443} (\bibinfo{year}{2004}), \bibinfo{pages}{823--826}.
\newblock


\bibitem[Gaskin et~al\mbox{.}(2012)]%
        {gaskin2012residential}
\bibfield{author}{\bibinfo{person}{Darrell~J Gaskin},
  \bibinfo{person}{Gniesha~Y Dinwiddie}, \bibinfo{person}{Kitty~S Chan}, {and}
  \bibinfo{person}{Rachael McCleary}.} \bibinfo{year}{2012}\natexlab{}.
\newblock \showarticletitle{Residential segregation and disparities in health
  care services utilization}.
\newblock \bibinfo{journal}{\emph{Medical Care Research and Review}}
  \bibinfo{volume}{69}, \bibinfo{number}{2} (\bibinfo{year}{2012}),
  \bibinfo{pages}{158--175}.
\newblock


\bibitem[Gianfrancesco et~al\mbox{.}(2018)]%
        {gianfrancesco2018potential}
\bibfield{author}{\bibinfo{person}{Milena~A Gianfrancesco},
  \bibinfo{person}{Suzanne Tamang}, \bibinfo{person}{Jinoos Yazdany}, {and}
  \bibinfo{person}{Gabriela Schmajuk}.} \bibinfo{year}{2018}\natexlab{}.
\newblock \showarticletitle{Potential biases in machine learning algorithms
  using electronic health record data}.
\newblock \bibinfo{journal}{\emph{JAMA internal medicine}}
  \bibinfo{volume}{178}, \bibinfo{number}{11} (\bibinfo{year}{2018}),
  \bibinfo{pages}{1544--1547}.
\newblock


\bibitem[Hardt et~al\mbox{.}(2016)]%
        {hardt2016equality}
\bibfield{author}{\bibinfo{person}{Moritz Hardt}, \bibinfo{person}{Eric Price},
  {and} \bibinfo{person}{Nati Srebro}.} \bibinfo{year}{2016}\natexlab{}.
\newblock \showarticletitle{Equality of opportunity in supervised learning}.
\newblock \bibinfo{journal}{\emph{Advances in neural information processing
  systems}}  \bibinfo{volume}{29} (\bibinfo{year}{2016}).
\newblock


\bibitem[Haukoos and Lewis(2015)]%
        {haukoos2015propensity}
\bibfield{author}{\bibinfo{person}{Jason~S Haukoos} {and}
  \bibinfo{person}{Roger~J Lewis}.} \bibinfo{year}{2015}\natexlab{}.
\newblock \showarticletitle{The propensity score}.
\newblock \bibinfo{journal}{\emph{Jama}} \bibinfo{volume}{314},
  \bibinfo{number}{15} (\bibinfo{year}{2015}), \bibinfo{pages}{1637--1638}.
\newblock


\bibitem[Heidari et~al\mbox{.}(2019)]%
        {heidari2019moral}
\bibfield{author}{\bibinfo{person}{Hoda Heidari}, \bibinfo{person}{Michele
  Loi}, \bibinfo{person}{Krishna~P Gummadi}, {and} \bibinfo{person}{Andreas
  Krause}.} \bibinfo{year}{2019}\natexlab{}.
\newblock \showarticletitle{A moral framework for understanding fair ml through
  economic models of equality of opportunity}. In
  \bibinfo{booktitle}{\emph{Proceedings of the conference on fairness,
  accountability, and transparency}}. \bibinfo{pages}{181--190}.
\newblock


\bibitem[Jiang and Nachum(2020)]%
        {jiang2020identifying}
\bibfield{author}{\bibinfo{person}{Heinrich Jiang} {and} \bibinfo{person}{Ofir
  Nachum}.} \bibinfo{year}{2020}\natexlab{}.
\newblock \showarticletitle{Identifying and correcting label bias in machine
  learning}. In \bibinfo{booktitle}{\emph{International Conference on
  Artificial Intelligence and Statistics}}. PMLR, \bibinfo{pages}{702--712}.
\newblock


\bibitem[Jiang et~al\mbox{.}(2020)]%
        {jiang2020wasserstein}
\bibfield{author}{\bibinfo{person}{Ray Jiang}, \bibinfo{person}{Aldo
  Pacchiano}, \bibinfo{person}{Tom Stepleton}, \bibinfo{person}{Heinrich
  Jiang}, {and} \bibinfo{person}{Silvia Chiappa}.}
  \bibinfo{year}{2020}\natexlab{}.
\newblock \showarticletitle{Wasserstein fair classification}. In
  \bibinfo{booktitle}{\emph{Uncertainty in Artificial Intelligence}}. PMLR,
  \bibinfo{pages}{862--872}.
\newblock


\bibitem[Johnson et~al\mbox{.}(2016)]%
        {johnson2016mimic}
\bibfield{author}{\bibinfo{person}{Alistair~EW Johnson}, \bibinfo{person}{Tom~J
  Pollard}, \bibinfo{person}{Lu Shen}, \bibinfo{person}{H~Lehman Li-Wei},
  \bibinfo{person}{Mengling Feng}, \bibinfo{person}{Mohammad Ghassemi},
  \bibinfo{person}{Benjamin Moody}, \bibinfo{person}{Peter Szolovits},
  \bibinfo{person}{Leo~Anthony Celi}, {and} \bibinfo{person}{Roger~G Mark}.}
  \bibinfo{year}{2016}\natexlab{}.
\newblock \showarticletitle{MIMIC-III, a freely accessible critical care
  database}.
\newblock \bibinfo{journal}{\emph{Scientific data}} \bibinfo{volume}{3},
  \bibinfo{number}{1} (\bibinfo{year}{2016}), \bibinfo{pages}{1--9}.
\newblock


\bibitem[Kallus and Zhou(2018)]%
        {kallus2018residual}
\bibfield{author}{\bibinfo{person}{Nathan Kallus} {and} \bibinfo{person}{Angela
  Zhou}.} \bibinfo{year}{2018}\natexlab{}.
\newblock \showarticletitle{Residual unfairness in fair machine learning from
  prejudiced data}. In \bibinfo{booktitle}{\emph{International Conference on
  Machine Learning}}. PMLR, \bibinfo{pages}{2439--2448}.
\newblock


\bibitem[Kamiran and Calders(2010)]%
        {kamiran2010classification}
\bibfield{author}{\bibinfo{person}{Faisal Kamiran} {and} \bibinfo{person}{Toon
  Calders}.} \bibinfo{year}{2010}\natexlab{}.
\newblock \showarticletitle{Classification with no discrimination by
  preferential sampling}. In \bibinfo{booktitle}{\emph{Proc. 19th Machine
  Learning Conf. Belgium and The Netherlands}}. Citeseer,
  \bibinfo{pages}{1--6}.
\newblock


\bibitem[Kamiran and Calders(2012)]%
        {kamiran2012data}
\bibfield{author}{\bibinfo{person}{Faisal Kamiran} {and} \bibinfo{person}{Toon
  Calders}.} \bibinfo{year}{2012}\natexlab{}.
\newblock \showarticletitle{Data preprocessing techniques for classification
  without discrimination}.
\newblock \bibinfo{journal}{\emph{Knowledge and Information Systems}}
  \bibinfo{volume}{33}, \bibinfo{number}{1} (\bibinfo{year}{2012}),
  \bibinfo{pages}{1--33}.
\newblock


\bibitem[Kilbertus et~al\mbox{.}(2020)]%
        {kilbertus2020sensitivity}
\bibfield{author}{\bibinfo{person}{Niki Kilbertus}, \bibinfo{person}{Philip~J
  Ball}, \bibinfo{person}{Matt~J Kusner}, \bibinfo{person}{Adrian Weller},
  {and} \bibinfo{person}{Ricardo Silva}.} \bibinfo{year}{2020}\natexlab{}.
\newblock \showarticletitle{The sensitivity of counterfactual fairness to
  unmeasured confounding}. In \bibinfo{booktitle}{\emph{Uncertainty in
  artificial intelligence}}. PMLR, \bibinfo{pages}{616--626}.
\newblock


\bibitem[Kingma and Welling(2013)]%
        {kingma2013auto}
\bibfield{author}{\bibinfo{person}{Diederik~P Kingma} {and}
  \bibinfo{person}{Max Welling}.} \bibinfo{year}{2013}\natexlab{}.
\newblock \showarticletitle{Auto-encoding variational bayes}.
\newblock \bibinfo{journal}{\emph{arXiv preprint arXiv:1312.6114}}
  (\bibinfo{year}{2013}).
\newblock


\bibitem[Krasanakis et~al\mbox{.}(2018)]%
        {krasanakis2018adaptive}
\bibfield{author}{\bibinfo{person}{Emmanouil Krasanakis},
  \bibinfo{person}{Eleftherios Spyromitros-Xioufis}, \bibinfo{person}{Symeon
  Papadopoulos}, {and} \bibinfo{person}{Yiannis Kompatsiaris}.}
  \bibinfo{year}{2018}\natexlab{}.
\newblock \showarticletitle{Adaptive sensitive reweighting to mitigate bias in
  fairness-aware classification}. In \bibinfo{booktitle}{\emph{Proceedings of
  the 2018 World Wide Web Conference}}. \bibinfo{pages}{853--862}.
\newblock


\bibitem[Kusner et~al\mbox{.}(2017)]%
        {kusner2017counterfactual}
\bibfield{author}{\bibinfo{person}{Matt~J Kusner}, \bibinfo{person}{Joshua
  Loftus}, \bibinfo{person}{Chris Russell}, {and} \bibinfo{person}{Ricardo
  Silva}.} \bibinfo{year}{2017}\natexlab{}.
\newblock \showarticletitle{Counterfactual fairness}.
\newblock \bibinfo{journal}{\emph{Advances in neural information processing
  systems}}  \bibinfo{volume}{30} (\bibinfo{year}{2017}).
\newblock


\bibitem[Larrazabal et~al\mbox{.}(2020)]%
        {larrazabal2020gender}
\bibfield{author}{\bibinfo{person}{Agostina~J Larrazabal},
  \bibinfo{person}{Nicol{\'a}s Nieto}, \bibinfo{person}{Victoria Peterson},
  \bibinfo{person}{Diego~H Milone}, {and} \bibinfo{person}{Enzo Ferrante}.}
  \bibinfo{year}{2020}\natexlab{}.
\newblock \showarticletitle{Gender imbalance in medical imaging datasets
  produces biased classifiers for computer-aided diagnosis}.
\newblock \bibinfo{journal}{\emph{Proceedings of the National Academy of
  Sciences}} \bibinfo{volume}{117}, \bibinfo{number}{23}
  (\bibinfo{year}{2020}), \bibinfo{pages}{12592--12594}.
\newblock


\bibitem[Louizos et~al\mbox{.}(2015)]%
        {louizos2015variational}
\bibfield{author}{\bibinfo{person}{Christos Louizos}, \bibinfo{person}{Kevin
  Swersky}, \bibinfo{person}{Yujia Li}, \bibinfo{person}{Max Welling}, {and}
  \bibinfo{person}{Richard Zemel}.} \bibinfo{year}{2015}\natexlab{}.
\newblock \showarticletitle{The variational fair autoencoder}.
\newblock \bibinfo{journal}{\emph{arXiv preprint arXiv:1511.00830}}
  (\bibinfo{year}{2015}).
\newblock


\bibitem[Ma et~al\mbox{.}(2018)]%
        {ma2018risk}
\bibfield{author}{\bibinfo{person}{Fenglong Ma}, \bibinfo{person}{Jing Gao},
  \bibinfo{person}{Qiuling Suo}, \bibinfo{person}{Quanzeng You},
  \bibinfo{person}{Jing Zhou}, {and} \bibinfo{person}{Aidong Zhang}.}
  \bibinfo{year}{2018}\natexlab{}.
\newblock \showarticletitle{Risk prediction on electronic health records with
  prior medical knowledge}. In \bibinfo{booktitle}{\emph{Proceedings of the
  24th ACM SIGKDD International Conference on Knowledge Discovery \& Data
  Mining}}. \bibinfo{pages}{1910--1919}.
\newblock


\bibitem[Mhasawade et~al\mbox{.}(2021)]%
        {mhasawade2021machine}
\bibfield{author}{\bibinfo{person}{Vishwali Mhasawade}, \bibinfo{person}{Yuan
  Zhao}, {and} \bibinfo{person}{Rumi Chunara}.}
  \bibinfo{year}{2021}\natexlab{}.
\newblock \showarticletitle{Machine learning and algorithmic fairness in public
  and population health}.
\newblock \bibinfo{journal}{\emph{Nature Machine Intelligence}}
  \bibinfo{volume}{3}, \bibinfo{number}{8} (\bibinfo{year}{2021}),
  \bibinfo{pages}{659--666}.
\newblock


\bibitem[Obermeyer et~al\mbox{.}(2019)]%
        {obermeyer2019dissecting}
\bibfield{author}{\bibinfo{person}{Ziad Obermeyer}, \bibinfo{person}{Brian
  Powers}, \bibinfo{person}{Christine Vogeli}, {and} \bibinfo{person}{Sendhil
  Mullainathan}.} \bibinfo{year}{2019}\natexlab{}.
\newblock \showarticletitle{Dissecting racial bias in an algorithm used to
  manage the health of populations}.
\newblock \bibinfo{journal}{\emph{Science}} \bibinfo{volume}{366},
  \bibinfo{number}{6464} (\bibinfo{year}{2019}), \bibinfo{pages}{447--453}.
\newblock


\bibitem[Pfohl et~al\mbox{.}(2019)]%
        {pfohl2019counterfactual}
\bibfield{author}{\bibinfo{person}{Stephen~R Pfohl}, \bibinfo{person}{Tony
  Duan}, \bibinfo{person}{Daisy~Yi Ding}, {and} \bibinfo{person}{Nigam~H
  Shah}.} \bibinfo{year}{2019}\natexlab{}.
\newblock \showarticletitle{Counterfactual reasoning for fair clinical risk
  prediction}. In \bibinfo{booktitle}{\emph{Machine Learning for Healthcare
  Conference}}. PMLR, \bibinfo{pages}{325--358}.
\newblock


\bibitem[Pfohl et~al\mbox{.}(2021)]%
        {pfohl2021empirical}
\bibfield{author}{\bibinfo{person}{Stephen~R Pfohl}, \bibinfo{person}{Agata
  Foryciarz}, {and} \bibinfo{person}{Nigam~H Shah}.}
  \bibinfo{year}{2021}\natexlab{}.
\newblock \showarticletitle{An empirical characterization of fair machine
  learning for clinical risk prediction}.
\newblock \bibinfo{journal}{\emph{Journal of biomedical informatics}}
  \bibinfo{volume}{113} (\bibinfo{year}{2021}), \bibinfo{pages}{103621}.
\newblock


\bibitem[Pleiss et~al\mbox{.}(2017)]%
        {pleiss2017fairness}
\bibfield{author}{\bibinfo{person}{Geoff Pleiss}, \bibinfo{person}{Manish
  Raghavan}, \bibinfo{person}{Felix Wu}, \bibinfo{person}{Jon Kleinberg}, {and}
  \bibinfo{person}{Kilian~Q Weinberger}.} \bibinfo{year}{2017}\natexlab{}.
\newblock \showarticletitle{On fairness and calibration}.
\newblock \bibinfo{journal}{\emph{arXiv preprint arXiv:1709.02012}}
  (\bibinfo{year}{2017}).
\newblock


\bibitem[Rajkomar et~al\mbox{.}(2018a)]%
        {rajkomar2018ensuring}
\bibfield{author}{\bibinfo{person}{Alvin Rajkomar}, \bibinfo{person}{Michaela
  Hardt}, \bibinfo{person}{Michael~D Howell}, \bibinfo{person}{Greg Corrado},
  {and} \bibinfo{person}{Marshall~H Chin}.} \bibinfo{year}{2018}\natexlab{a}.
\newblock \showarticletitle{Ensuring fairness in machine learning to advance
  health equity}.
\newblock \bibinfo{journal}{\emph{Annals of internal medicine}}
  \bibinfo{volume}{169}, \bibinfo{number}{12} (\bibinfo{year}{2018}),
  \bibinfo{pages}{866--872}.
\newblock


\bibitem[Rajkomar et~al\mbox{.}(2018b)]%
        {rajkomar2018scalable}
\bibfield{author}{\bibinfo{person}{Alvin Rajkomar}, \bibinfo{person}{Eyal
  Oren}, \bibinfo{person}{Kai Chen}, \bibinfo{person}{Andrew~M Dai},
  \bibinfo{person}{Nissan Hajaj}, \bibinfo{person}{Michaela Hardt},
  \bibinfo{person}{Peter~J Liu}, \bibinfo{person}{Xiaobing Liu},
  \bibinfo{person}{Jake Marcus}, \bibinfo{person}{Mimi Sun}, {et~al\mbox{.}}}
  \bibinfo{year}{2018}\natexlab{b}.
\newblock \showarticletitle{Scalable and accurate deep learning with electronic
  health records}.
\newblock \bibinfo{journal}{\emph{NPJ Digital Medicine}} \bibinfo{volume}{1},
  \bibinfo{number}{1} (\bibinfo{year}{2018}), \bibinfo{pages}{1--10}.
\newblock


\bibitem[Rezaei et~al\mbox{.}(2021)]%
        {rezaei2021robust}
\bibfield{author}{\bibinfo{person}{Ashkan Rezaei}, \bibinfo{person}{Anqi Liu},
  \bibinfo{person}{Omid Memarrast}, {and} \bibinfo{person}{Brian~D Ziebart}.}
  \bibinfo{year}{2021}\natexlab{}.
\newblock \showarticletitle{Robust fairness under covariate shift}. In
  \bibinfo{booktitle}{\emph{Proceedings of the AAAI Conference on Artificial
  Intelligence}}, Vol.~\bibinfo{volume}{35}. \bibinfo{pages}{9419--9427}.
\newblock


\bibitem[Sattigeri et~al\mbox{.}(2019)]%
        {sattigeri2019fairness}
\bibfield{author}{\bibinfo{person}{Prasanna Sattigeri},
  \bibinfo{person}{Samuel~C Hoffman}, \bibinfo{person}{Vijil Chenthamarakshan},
  {and} \bibinfo{person}{Kush~R Varshney}.} \bibinfo{year}{2019}\natexlab{}.
\newblock \showarticletitle{Fairness GAN: Generating datasets with fairness
  properties using a generative adversarial network}.
\newblock \bibinfo{journal}{\emph{IBM Journal of Research and Development}}
  \bibinfo{volume}{63}, \bibinfo{number}{4/5} (\bibinfo{year}{2019}),
  \bibinfo{pages}{3--1}.
\newblock


\bibitem[Seyyed-Kalantari et~al\mbox{.}(2020)]%
        {seyyed2020chexclusion}
\bibfield{author}{\bibinfo{person}{Laleh Seyyed-Kalantari},
  \bibinfo{person}{Guanxiong Liu}, \bibinfo{person}{Matthew McDermott},
  \bibinfo{person}{Irene~Y Chen}, {and} \bibinfo{person}{Marzyeh Ghassemi}.}
  \bibinfo{year}{2020}\natexlab{}.
\newblock \showarticletitle{CheXclusion: Fairness gaps in deep chest X-ray
  classifiers}. In \bibinfo{booktitle}{\emph{BIOCOMPUTING 2021: Proceedings of
  the Pacific Symposium}}. World Scientific, \bibinfo{pages}{232--243}.
\newblock


\bibitem[Sohn et~al\mbox{.}(2015)]%
        {sohn2015learning}
\bibfield{author}{\bibinfo{person}{Kihyuk Sohn}, \bibinfo{person}{Honglak Lee},
  {and} \bibinfo{person}{Xinchen Yan}.} \bibinfo{year}{2015}\natexlab{}.
\newblock \showarticletitle{Learning structured output representation using
  deep conditional generative models}.
\newblock \bibinfo{journal}{\emph{Advances in neural information processing
  systems}}  \bibinfo{volume}{28} (\bibinfo{year}{2015}),
  \bibinfo{pages}{3483--3491}.
\newblock


\bibitem[Veinot et~al\mbox{.}(2018)]%
        {veinot2018good}
\bibfield{author}{\bibinfo{person}{Tiffany~C Veinot}, \bibinfo{person}{Hannah
  Mitchell}, {and} \bibinfo{person}{Jessica~S Ancker}.}
  \bibinfo{year}{2018}\natexlab{}.
\newblock \showarticletitle{Good intentions are not enough: how informatics
  interventions can worsen inequality}.
\newblock \bibinfo{journal}{\emph{Journal of the American Medical Informatics
  Association}} \bibinfo{volume}{25}, \bibinfo{number}{8}
  (\bibinfo{year}{2018}), \bibinfo{pages}{1080--1088}.
\newblock


\bibitem[Wang and Blei(2019)]%
        {wang2019blessings}
\bibfield{author}{\bibinfo{person}{Yixin Wang} {and} \bibinfo{person}{David~M
  Blei}.} \bibinfo{year}{2019}\natexlab{}.
\newblock \showarticletitle{The blessings of multiple causes}.
\newblock \bibinfo{journal}{\emph{J. Amer. Statist. Assoc.}}
  \bibinfo{volume}{114}, \bibinfo{number}{528} (\bibinfo{year}{2019}),
  \bibinfo{pages}{1574--1596}.
\newblock


\bibitem[Wang and Blei(2020)]%
        {wang2020towards}
\bibfield{author}{\bibinfo{person}{Yixin Wang} {and} \bibinfo{person}{David~M
  Blei}.} \bibinfo{year}{2020}\natexlab{}.
\newblock \showarticletitle{Towards clarifying the theory of the deconfounder}.
\newblock \bibinfo{journal}{\emph{arXiv preprint arXiv:2003.04948}}
  (\bibinfo{year}{2020}).
\newblock


\bibitem[Wang et~al\mbox{.}(2020)]%
        {wang2020causal}
\bibfield{author}{\bibinfo{person}{Yixin Wang}, \bibinfo{person}{Dawen Liang},
  \bibinfo{person}{Laurent Charlin}, {and} \bibinfo{person}{David~M Blei}.}
  \bibinfo{year}{2020}\natexlab{}.
\newblock \showarticletitle{Causal inference for recommender systems}. In
  \bibinfo{booktitle}{\emph{Fourteenth ACM Conference on Recommender Systems}}.
  \bibinfo{pages}{426--431}.
\newblock


\bibitem[Zemel et~al\mbox{.}(2013)]%
        {zemel2013learning}
\bibfield{author}{\bibinfo{person}{Rich Zemel}, \bibinfo{person}{Yu Wu},
  \bibinfo{person}{Kevin Swersky}, \bibinfo{person}{Toni Pitassi}, {and}
  \bibinfo{person}{Cynthia Dwork}.} \bibinfo{year}{2013}\natexlab{}.
\newblock \showarticletitle{Learning fair representations}. In
  \bibinfo{booktitle}{\emph{International conference on machine learning}}.
  PMLR, \bibinfo{pages}{325--333}.
\newblock


\bibitem[Zhang et~al\mbox{.}(2019)]%
        {zhang2019medical}
\bibfield{author}{\bibinfo{person}{Linying Zhang}, \bibinfo{person}{Yixin
  Wang}, \bibinfo{person}{Anna Ostropolets}, \bibinfo{person}{Jami~J Mulgrave},
  \bibinfo{person}{David~M Blei}, {and} \bibinfo{person}{George Hripcsak}.}
  \bibinfo{year}{2019}\natexlab{}.
\newblock \showarticletitle{The medical deconfounder: assessing treatment
  effects with electronic health records}. In \bibinfo{booktitle}{\emph{Machine
  Learning for Healthcare Conference}}. PMLR, \bibinfo{pages}{490--512}.
\newblock


\bibitem[Zhang et~al\mbox{.}(2017)]%
        {zhang2017leap}
\bibfield{author}{\bibinfo{person}{Yutao Zhang}, \bibinfo{person}{Robert Chen},
  \bibinfo{person}{Jie Tang}, \bibinfo{person}{Walter~F Stewart}, {and}
  \bibinfo{person}{Jimeng Sun}.} \bibinfo{year}{2017}\natexlab{}.
\newblock \showarticletitle{LEAP: learning to prescribe effective and safe
  treatment combinations for multimorbidity}. In
  \bibinfo{booktitle}{\emph{proceedings of the 23rd ACM SIGKDD international
  conference on knowledge Discovery and data Mining}}.
  \bibinfo{pages}{1315--1324}.
\newblock


\bibitem[Zhou et~al\mbox{.}(2021)]%
        {zhou2021radfusion}
\bibfield{author}{\bibinfo{person}{Yuyin Zhou}, \bibinfo{person}{Shih-Cheng
  Huang}, \bibinfo{person}{Jason~Alan Fries}, \bibinfo{person}{Alaa Youssef},
  \bibinfo{person}{Timothy~J Amrhein}, \bibinfo{person}{Marcello Chang},
  \bibinfo{person}{Imon Banerjee}, \bibinfo{person}{Daniel Rubin},
  \bibinfo{person}{Lei Xing}, \bibinfo{person}{Nigam Shah}, {et~al\mbox{.}}}
  \bibinfo{year}{2021}\natexlab{}.
\newblock \showarticletitle{RadFusion: Benchmarking Performance and Fairness
  for Multimodal Pulmonary Embolism Detection from CT and EHR}.
\newblock \bibinfo{journal}{\emph{arXiv preprint arXiv:2111.11665}}
  (\bibinfo{year}{2021}).
\newblock


\end{thebibliography}
